\documentclass{article}

\usepackage{microtype}
\usepackage{subcaption}
\usepackage{booktabs} 
\usepackage{blkarray}
\usepackage{latexsym}
\usepackage{amsmath}
\usepackage{amssymb}
\usepackage{amsthm}
\usepackage{dcolumn,booktabs}
\usepackage{epsfig}
\usepackage{enumitem}
\usepackage{graphicx}
\usepackage{mathrsfs}
\usepackage{natbib}
\usepackage{float}

\usepackage{hyperref}

\newcommand{\R}{\mathbb{R}}

\newcommand{\N}{\mathbb{N}}

\newcommand{\ol}{\overline}
\newcommand{\mbf}{\mathbf}

\newcommand{\mscr}{\mathscr}

\newcommand{\eval}{\bigr\rvert}

\newcommand{\coef}{\text{\normalfont coef}}
\newcommand{\asgn}{\text{\normalfont asgn}}
\newcommand{\comp}{\text{\normalfont comp}}
\newcommand{\diag}{\text{\normalfont diag}}
\newcommand\widebar[1]{\mathop{\overline{#1}}}

\newcolumntype{d}[1]{D{.}{.}{#1}}

\theoremstyle{plain} 
\newtheorem{thm}{Theorem}
\newtheorem{cor}{Corollary}
\newtheorem{lem}{Lemma}
\newtheorem{prop}{Proposition}

\theoremstyle{definition}

\newtheorem{defn}{Definition}

\theoremstyle{remark}

\usepackage[accepted]{icml2021}

\icmltitlerunning{Probabilistic Generating Circuits}

\begin{document}

\twocolumn[
\icmltitle{Probabilistic Generating Circuits}



\icmlsetsymbol{equal}{*}

\begin{icmlauthorlist}
\icmlauthor{Honghua Zhang}{ucla}
\icmlauthor{Brendan Juba}{wustl}
\icmlauthor{Guy Van den Broeck}{ucla}
\end{icmlauthorlist}

\icmlaffiliation{ucla}{Computer Science Department, University of California Los Angeles, USA}
\icmlaffiliation{wustl}{Computer Science Department, Washington University in St. Louis, Missouri, USA}

\icmlcorrespondingauthor{Honghua Zhang}{hzhang19@cs.ucla.edu}

\icmlkeywords{Machine Learning, ICML}

\vskip 0.3in
]



\printAffiliationsAndNotice{}  

\begin{abstract}
Generating functions, which are widely used in combinatorics and probability theory, encode function values into the coefficients of a polynomial. In this paper, we explore their use as a tractable probabilistic model, and propose probabilistic generating circuits (PGCs) for their efficient representation. PGCs are strictly more expressive efficient than many existing tractable probabilistic models, including determinantal point processes (DPPs), probabilistic circuits (PCs) such as sum-product networks, and tractable graphical models. We contend that PGCs are not just a theoretical framework that unifies vastly different existing models, but also show great potential in modeling realistic data. We exhibit a simple class of PGCs that are not trivially subsumed by simple combinations of PCs and DPPs, and obtain competitive performance on a suite of density estimation benchmarks. We also highlight PGCs' connection to the theory of strongly Rayleigh distributions.
\end{abstract}

\section{Introduction}
Probabilistic modeling is an important task in machine learning.
Scaling up such models is a
key challenge: probabilistic inference quickly
becomes intractable as the models become large and sophisticated~\cite{roth1996hardness}.
Central to this effort is the development of
\emph{tractable probabilistic models} (TPMs) that guarantee tractable
probabilistic inference in the size of the model, yet can efficiently represent a wide range of
probability distributions. There has been a proliferation of different classes of TPMs.
Examples include bounded-treewidth graphical models~\citep{meila2000learning},
determinantal point processes~\citep{borodin2005eynard,MAL-044},
and various probabilistic circuits~\citep{darwiche2009modeling,kisa2014probabilistic,AAAI-Tutorial}
such as sum-product~networks~\citep{poon2011sum}.


Ideally, we want our probabilistic models to be as
\emph{expressive efficient}~\citep{martens2014expressive} as possible, meaning
that they can \emph{efficiently} represent as many classes of distributions as
possible, and adapt to a wider spectrum of realistic applications.
Often, however, stronger expressive power comes at the expense of tractability:
fewer restrictions can make a model more expressive efficient, but it can also make
probabilistic inference intractable.
We therefore raise the following central research question of this paper:
\emph{Does there exist a class of tractable probabilistic models that is
strictly more expressive efficient than current~TPMs?}


All aforementioned models are usually seen as representing \emph{probability mass functions}:
they take assignments to random variables as input and output likelihoods.
In contrast, especially in the field of probability theory,
it is also common to represent distributions as
\emph{probability generating polynomials} (or generating polynomials
for short). Generating polynomials are a powerful mathematical tool,
but they have not yet found direct use as a probabilistic machine learning representation
that permits tractable probabilistic~inference.

We make the key observation that the marginal probabilities
(including likelihoods) for a probability distribution can
be computed by evaluating its generating
polynomial in a particular way. Based on this observation, we propose
\emph{probabilistic generating circuits} (PGCs), a class of
probabilistic models that represent probability generating polynomials
compactly as directed acyclic graphs.
PGCs provide a partly positive answer to our research question: they are
the first known class of TPMs that are strictly more 
expressive efficient than decomposable probabilistic circuits
(PCs), in particular, sum-product networks, and determinantal point processes (DPPs) while supporting tractable marginal inference.

\begin{figure*}
\centering
  \begin{subfigure}[b]{0.22\linewidth}
  \centering
  {\footnotesize
  \begin{tabular}{c c c|c}
  \hline
  $X_1$ & $X_2$ & $X_3$ & ${\Pr}_{\beta}$ \\ \hline
  $0$ & $0$ & $0$ & $0.02$ \\
  $0$ & $0$ & $1$ & $0.08$ \\
  $0$ & $1$ & $0$ & $0.12$ \\
  $0$ & $1$ & $1$ & $0.48$ \\
  $1$ & $0$ & $0$ & $0.02$ \\
  $1$ & $0$ & $1$ & $0.08$ \\
  $1$ & $1$ & $0$ & $0.04$ \\
  $1$ & $1$ & $1$ & $0.16$ \\ \hline
  \end{tabular}}
  \caption{Table}
  \label{fig:example_dist_table}
  \end{subfigure}
  ~
  \begin{subfigure}[b]{0.26\linewidth}
  \centering
  \includegraphics[height=4.0cm]{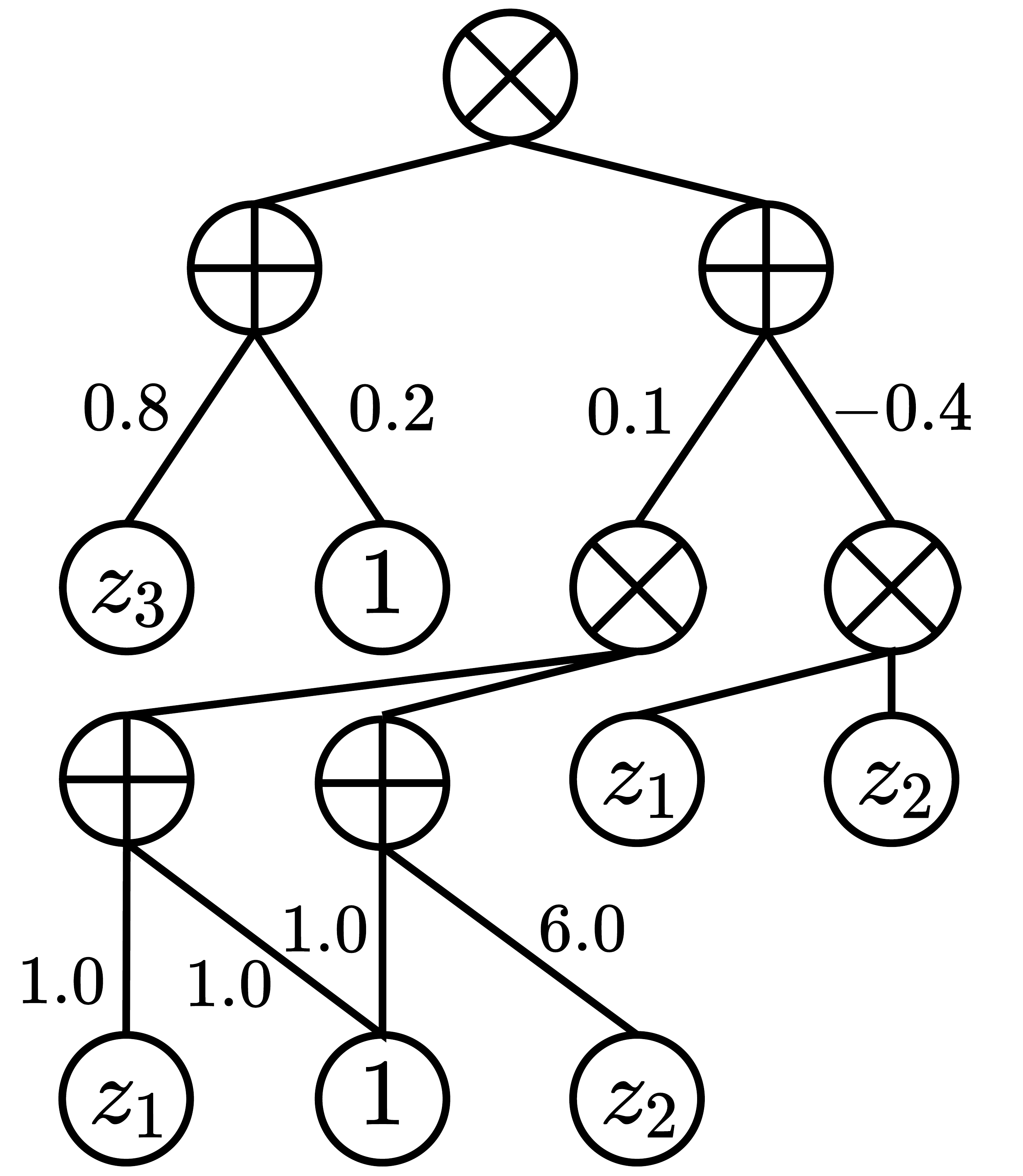}
  \caption{Probabilistic generating circuit}
  \label{fig:example_dist_pgc}
  \end{subfigure}
  ~
  \begin{subfigure}[b]{0.22\linewidth}
  \centering
  \includegraphics[height=4.0cm]{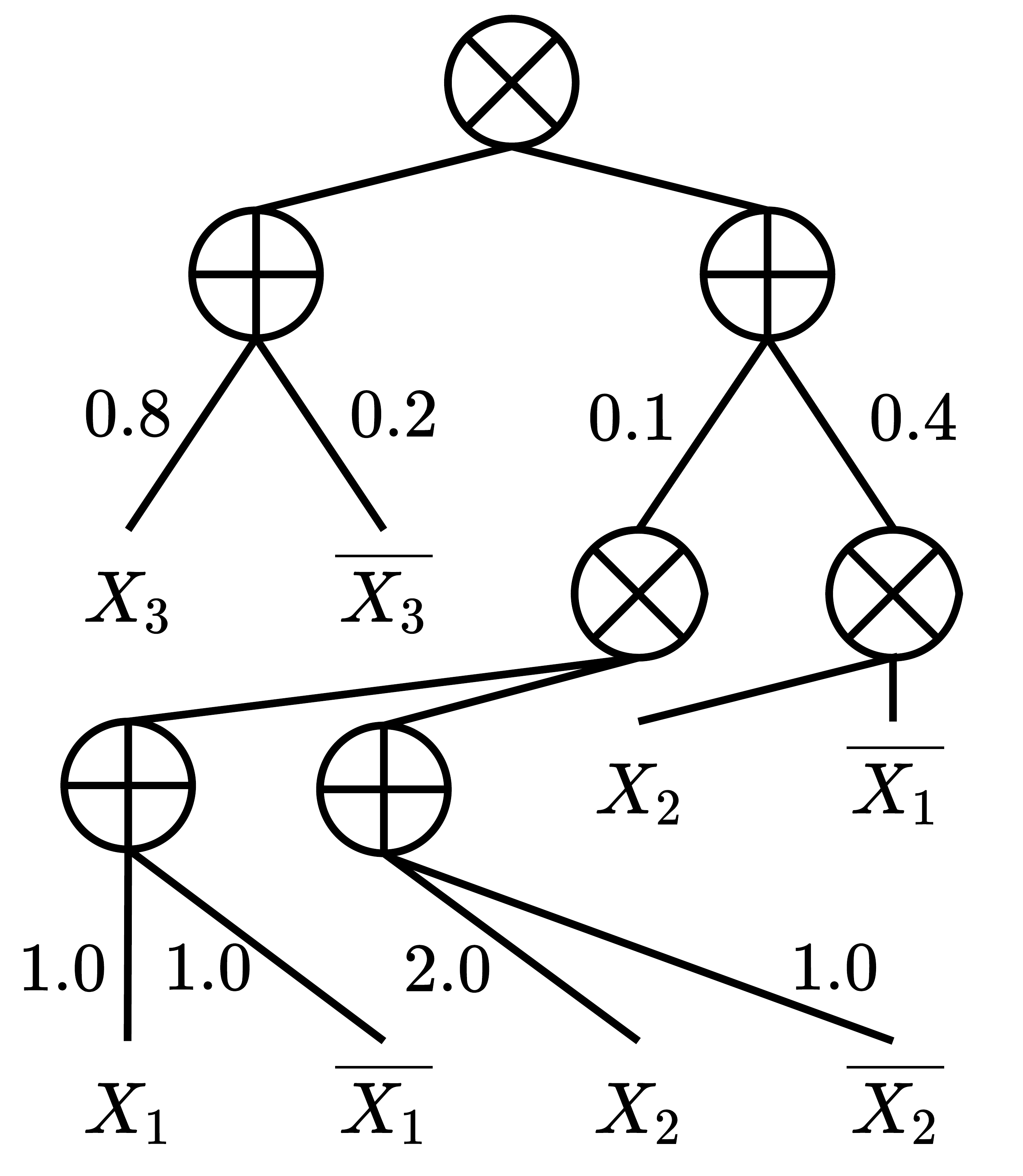}
  \caption{Probabilistic mass circuit}
  \label{fig:example_dist_pc}
  \end{subfigure}
  ~
  \begin{subfigure}[b]{0.22\linewidth}
      {\footnotesize
      \begin{align*}
         L^{\beta} &= \begin{blockarray}{cccc}
                X_1 & X_2 & X_3  \\
                \begin{block}{[ccc]c}
                1 & 2 & 0 & X_1 \\
                2 & 6 & 0 & X_2 \\
                0 & 0 & 4 & X_3 \\
                \end{block}
              \end{blockarray} \\
         K^{\beta} &= \begin{blockarray}{cccc}
                X_1 & X_2 & X_3 \\
                \begin{block}{[ccc]c}
                0.3 & 0.2 & 0 & X_1 \\
                0.2 & 0.8 & 0 & X_2 \\
                0 & 0 & 0.8 & X_3 \\
                \end{block}
              \end{blockarray}
     \end{align*}}
  \vspace{-20pt}
  \caption{Kernel $L^{\beta}$ and marginal kernel $K^{\beta}$
  for a DPP}
  \label{fig:example_dist_dpp}
  \end{subfigure}
\caption{Four different representations for the same probability
  distribution ${\Pr}_{\beta}$.}
\end{figure*}

Section~\ref{sec:pgc} formally defines PGCs and establishes their
tractability by presenting an efficient algorithm for computing marginals.
Section~\ref{sec:pgc_subsume_tpm} demonstrates the expressive power of PGCs by
showing that they subsume PCs and DPPs while remaining
strictly more expressive efficient.
Section~\ref{sec:beyond_pc_dpp} shows that there are PGCs
that cannot be represented by PCs with DPPs as leaves.
Section~\ref{sec:experiments} evaluates PGCs on
standard density estimation benchmarks: even the simplest
PGCs outperform other TPM learners on half of the datasets.
Then, Section~\ref{sec:pgc_sr_distribution} highlights PGCs' connection
to strongly Rayleigh distributions.
Section~\ref{sec:conclusion} summarizes the paper and motivates future research~directions.

\section{Probabilistic Generating Circuits}
\label{sec:pgc}
In this section we establish probabilistic generating circuits (PGCs) as
a class of tractable probabilistic models.
We first introduce generating polynomials as a representation for
probability distributions and propose PGCs for their compact representations.
Then, we show that marginal probabilities for a PGC
can be computed efficiently.

\subsection{Probability Generating Polynomials}
\label{sec:pgc_generating_polynomials}
It is a common technique in combinatorics to encode
sequences as \emph{generating polynomials}.
In particular, probability distributions over binary random
variables can be represented by \emph{probability generating polynomials}.

\begin{defn}
Let $\Pr(\cdot)$ be a probability distribution over binary random variables $X_1, X_2, \dots,
X_n$, then the \emph{probability generating polynomial} (or generating polynomial for short)
for the distribution is defined as
\begin{align}
\label{eq:pgc_generating_polynomial}
g(z_1, \dots, z_n) = \sum_{S \subseteq \{1, \dots, n\}} \alpha_{S}z^{S}
\end{align}
where $\alpha_{S} = \Pr(\{X_i \!\!=\!\! 1\}_{i \in S}, \{X_i \!\!=\!\! 0\}_{i \notin S})$ and
$z^{S} = \prod_{i \in S} z_{i}$.
\end{defn}

As an illustrating example, we consider the probability distribution $\Pr_{\beta}$ specified
as a table in
Figure~\ref{fig:example_dist_table}.
By definition, the generating polynomial for
distribution $\Pr_{\beta}$ is given by
\vspace{-4pt}
\begin{align}
\begin{split}
\label{eq:generating_polynomial_plain}
g_{\beta} &= 0.16z_1z_2z_3 + 0.04z_1z_2 + 0.08z_1z_3 + 0.02z_1\\
       &\quad + 0.48z_2z_3 + 0.12z_2 + 0.08z_3 + 0.02.
\end{split}
\end{align}
We see from Equation~\ref{eq:generating_polynomial_plain} that the generating
polynomial for a distribution simply ``enumerates'' all possible variable assignments term-by-term,
and the coefficient of each term corresponds to the probability of an assignment.
The probability for the assignment $X_1 = 0, X_2 = 1, X_3 = 1$, for example, is $0.48$,
which corresponds to the coefficient of the term $z_2z_3$.
That is, given an assignment, we can evaluate its probability
by directly reading off the coefficient for the corresponding term.
We can also evaluate marginal probabilities by summing over the coefficients
for a \emph{set} of terms.
For example, the marginal probability $\Pr(X_2 = 0, X_3 = 0)$ is given by
$\Pr(X_1 = 0, X_2 = 0, X_3 = 0) + \Pr(X_1 = 1, X_2 = 0, X_3 = 0)$, which corresponds to
the sum of the constant term and the coefficient for the term $z_1$.

\subsection{Compactly Representing Generating Polynomials}
Equation~\ref{eq:generating_polynomial_plain} also illustrates
that the size of a term-by-term representation
for a generating polynomial is \emph{exponential} in the number of
variables. As the number of variables increases,
it quickly becomes impractical to compute probabilities by
extracting coefficients from these polynomials.
Hence, to turn generating polynomials into tractable models,
we need a data structure to represent them more efficiently.
We thus introduce
a new class of probabilistic circuits called \emph{probabilistic generating circuits}
to represent generating polynomials compactly as directed acyclic graphs (DAGs).
We first present the formal definition for PGCs.
\begin{defn}
A \emph{probabilistic generating circuit} (PGC) is a directed acyclic graph
consisting of three types of nodes:
\vspace{-5pt}
\begin{enumerate}[noitemsep]
\item \emph{Sum nodes} $\bigoplus$ with weighted edges to children;
\item \emph{Product nodes} $\bigotimes$ with unweighted edges to children;
\item \emph{Leaf nodes}, which are $z_i$ or constants.
\vspace{-4pt}
\end{enumerate}
A PGC has one node of out-degree $0$ (edges are directed from
children to parents),
and we refer to it as the \emph{root} of the PGC.
The \emph{size} of a PGC is the number of edges in it.
\end{defn}
Each node in a PGC \emph{represents} a polynomial,
(i) each leaf in a PGC represents the polynomial $z_i$ or a constant, (ii) each
sum node represents the weighted sum over the polynomials represented by its children, and
(iii) each product node represents the unweighted product over the polynomials represented
by its children. The polynomial represented by a PGC is the polynomial represented by its root.

We have now fully specified the \emph{syntax} of PGCs, but a PGC with valid syntax
does not necessarily have valid \emph{semantics}. Because of the presence of negative
parameters, it is not guaranteed that the polynomial represented by a PGC is a
probability generating polynomial: it might contain terms that are not multiaffine
or have negative coefficients (e.g. $-1.2z_1z_2^{3}$).
In practice, however, we show in Section~\ref{sec:beyond_pc_dpp} that, by
certain compositional operations, we can construct
PGCs that are guaranteed to have valid semantics for any parameterization.

Continuing our example, we observe that
the generating polynomial $g_{\beta}$ in Equation~\ref{eq:generating_polynomial_plain}
can be re-written as:
\vspace{-2pt}
\begin{align}
\label{eq:generating_polynomial_compact}
(0.1(z_1 + 1)(6z_2 + 1) - 0.4z_1z_2)(0.8z_3 + 0.2)
\end{align}
Based on Equation~\ref{eq:generating_polynomial_compact}, we can immediately construct
a PGC that compactly represents $g_{\beta}$, as shown in Figure~\ref{fig:example_dist_pgc}.
In this way, generating polynomials for high-dimensional distributions may become feasible to represent by PGCs.

\subsection{Tractable Inference with PGCs}
We now show that the computation of marginals is tractable for PGCs.
As briefly mentioned in Section~\ref{sec:pgc_generating_polynomials},
we can compute probabilities by extracting the coefficients of generating polynomials,
which is much trickier when they are represented as deeply-nested DAGs; as shown in
Figure~\ref{fig:example_dist_pgc}, it is impossible to directly read off any coefficient.
We circumvent this problem by making the key observation that we can
``zero-out'' the terms we don't want
from generating polynomials by evaluating them in a certain way. For example,
when evaluating a marginal probability with $X_1$ set to $0$, we zero-out all terms
that contain $z_1$ by setting $z_1$ to $0$. We generalize this idea as follows.
\begin{lem}
\label{lem:pgc_general_marginal}
Let $g(z_1, \dots, z_n)$ be a probability generating polynomial for $\Pr(\cdot)$, then for
$A, B \subseteq \{1, \dots, n\}$ with $A \cap B = \emptyset$,
the marginal probability can be computed by:
\begin{align*}
  &\Pr(\{X_i = 1\}_{i \in A}, \{X_i = 0\}_{i \in B}) \\
&= {\coef}_{|A|} \left( g(\{z_i = t\}_{i \in A}, \{z_i = 0\}_{i \in B}, \{z_i = 1\}_{i \notin {A \cup B}}) \right),
\end{align*}
where $t$ is an indeterminate for polynomials, and ${\coef}_{k}(g(t))$ denotes the coefficient
for the term $t^{k}$ in $g(t)$.
\end{lem}
Lemma~\ref{lem:pgc_general_marginal} basically says that for a generating polynomial,
its marginals can be computed by evaluating the generating polynomial in the
\emph{polynomial ring} $\R[t]$. With a bottom-up pass, this result
naturally extends to generating polynomials represented as PGCs: we
first evaluate the leaf nodes to $t$, $1$ or $0$ based on the
assignment; then, for the sum nodes, we compute the weighted sum over the polynomials
that their children evaluate to; for the product nodes, we compute the product over
the polynomials that their children evaluate to.
Note that, as we are
taking sums and products over univariate polynomials of degree $n$,
the time complexities for the naive algorithms are $O(n)$ and $O(n^2)$, respectively.
In light of this, it is not hard to see that computing marginal probabilities
is polynomial-time with respect to the size of the PGCs.

\begin{thm}
For PGCs of size $m$ representing distributions on $n$ binary
random variables, marginal probabilities (including
likelihoods) are computable in time $O(mn\log n\log\log n)$.
\end{thm}

The $O(mn\log n\log\log n)$ complexity in the theorem consists of two parts:
the $O(m)$ part
is the time complexity of a bottom-up pass for a PGC of size $m$
and the $O(n\log n\log\log n)$ part is contributed
by the time complexity of computing the product of two degree-$n$ polynomials with
fast Fourier transform~\citep{schonhage1971schnelle,cantor1991fast}.

\section{PGCs Subsume Other Probabilistic Models}
\label{sec:pgc_subsume_tpm}
To this point,
we have introduced PGCs as a probabilistic model
and shown that they support tractable marginals.
Next, we show that PGCs are \emph{strictly} more expressive efficient than other TPMs by 
showing that PGCs tractably subsume decomposable probabilistic circuits and 
determinantal point processes.

\subsection{PGCs Subsume Other Probabilistic Circuits}
\label{sec:pgc_subsume_pc}
We start by introducing the basics of probabilistic
circuits~\citep{AAAI-Tutorial, LecNoAAAI20}.
Just like PGCs, each PC also represents a polynomial with respect to variables
$X_i$ and $\ol{X_i}$.
The syntax of probabilistic circuits (PCs)
is basically the same as PGCs except for the following:
\begin{enumerate}
\item the variables in PCs are $X_i$s and $\widebar{X_i}$s rather than $z_i$s; they
are inherently different in the sense that $X_i$s and $\widebar{X_i}$s are the random
variables themselves, while $z_i$s are symbolic formal objects;
\item the edge weights of PCs must be non-negative;
\item unlike PGCs, which represent probability generating polynomials,
all of the existing PCs represent probability mass functions (as polynomials), so we
sometimes refer to them as \emph{probabilistic mass circuits}.
\vspace{-4pt}
\end{enumerate}
Figure~\ref{fig:example_dist_pc} shows an example PC that represents the
distribution ${\Pr}_{\beta}$.
For a given assignment $\mbf{X}\!=\!\mbf{x}$,
the PC~$\mscr{A}$ \emph{evaluates} to a number $\mscr{A}(\mbf{x})$,
which is obtained by (i)~replacing $X_i$ variable leaves by $x_i$, (ii)~replacing $\widebar{X_i}$ variable
leaves by $1-x_i$, (iii)~evaluating product nodes as taking
the product over their children, and (iv)~evaluating sum nodes as taking
a weighted sum over their children.
Finally, a PC $\mscr{A}$ with variable leaves $\mbf{X} = (X_1, \dots, X_n)$
represents the probability distribution
$\Pr(\mbf{X}\!=\!\mbf{x}) \propto \mscr{A}(\mbf{x})$. 

For an arbitrary PC, most
probabilistic inference tasks, including marginals and MAP inference,
are computationally hard in the circuit size.
In order to guarantee the efficient evaluation of queries it is
therefore necessary to impose further constraints on the structure of the
circuit. In this paper we consider two well-known structural properties of probabilistic
circuits~\citep{darwiche2002knowledge,LecNoAAAI20}:
\begin{defn}
For a PC, we denote the input variables that a node
depends on as its \emph{scope}; then,
\vspace{-5pt}
\begin{enumerate}[noitemsep]
\item A $\bigotimes$ node is \emph{decomposable} if the scopes of its children
    are disjoint.
\item A $\bigoplus$  node is \emph{smooth} if the scopes of its children
    are the same.
\end{enumerate}
\vspace{-4pt}
A PC is decomposable if all of its $\bigotimes$ nodes are decomposable;
a PC is smooth if all of its $\bigoplus$ nodes are smooth.
\end{defn}
Let $\mscr{A}$ be
a PC over $X_1, \dots, X_n$. If $\mscr{A}$ is decomposable and smooth,
then we can efficiently compute its marginals: for
disjoint $A, B \subseteq \{1, \dots, n\}$
the marginal probability $\Pr(\{X_i = 1\}_{i \in A}, \{X_i = 0\}_{i \in B})$
is given by the evaluation of $\mscr{A}$ with the following inputs.
$$
\begin{cases}
X_{i} = 1, \widebar{X_{i}} = 0 & \text{ if } i \in A \\
X_{i} = 0, \widebar{X_{i}} = 1 & \text{ if } i \in B \\
X_{i} = 1, \widebar{X_{i}} = 1 & \text{otherwise.}
\end{cases}
$$
Many TPMs are certain forms of decomposable PCs.
Examples include sum-product networks (SPNs)~\citep{poon2011sum, peharz2019random},
And-Or graphs~\citep{mateescu2008and},
probabilistic sentential decision
diagrams (PSDDs)~\citep{kisa2014probabilistic}, arithmetic circuits~\citep{darwiche2009modeling}, cutset
networks~\citep{rahman2016learning} and bounded-treewidth graphical
models~\citep{meila2000learning} such as Chow-Liu trees~\citep{chow1968approximating} and
hidden Markov models~\citep{rabiner1986introduction}.

A decomposable PC can always be ``smoothed'' (i.e. transformed into a
smooth and decomposable PC) in
polynomial time with respect to its size~\citep{darwiche2001tractable,shih2019smoothing}.
Hence, when we are trying to show that decomposable PCs can be transformed
into equivalent PGCs in polynomial time, we can always assume without loss of
generality that decomposable PCs are also smooth.
Our first observation
is that the probability mass functions represented by smooth and decomposable PCs
are very similar to the corresponding generating polynomials:
\begin{prop}
\label{prop:pc_to_pgc}
Let $\mscr{A}$ be a smooth and decomposable PC that represents the probability distribution
$\Pr$ over random variables $X_1, \dots, X_n$. Then $\mscr{A}$ represents a probability mass
polynomial of the form:
\begin{align}
\label{eq:pc_mass_polynomial}
m(z_1, \dots, z_n) = \sum_{S \subseteq \{1, \dots, n\}} \alpha_{S}
\prod_{i \in S} X_i \prod_{i \notin S} \widebar{X_i}
\end{align}
where $\alpha_{S} = \Pr(\{X_i = 1\}_{i \in S}, \{X_i = 0\}_{i \notin S})$.
\end{prop}
Note that Equation~\ref{eq:pc_mass_polynomial} is closely related to
the network polynomials~\citep{darwiche2003differential} defined for Bayesian Networks.
By comparing Equation~\ref{eq:pc_mass_polynomial} to Equation~\ref{eq:pgc_generating_polynomial}
in the definition of generating circuits,
we find that they look very similar, except for the
absence of the negative random variables $\widebar{X_i}$ in
Equation~\ref{eq:pgc_generating_polynomial}, which gives us the following corollary:
\begin{cor}
\label{coro:pc_to_pgc}
Let $\mscr{A}$ be a smooth and decomposable PC. By replacing all
$\overline{X_{i}}$ in $\mscr{A}$ by $1$ and $X_i$ by $z_i$, we obtain a PGC that represents
the same distribution.
\end{cor}
Corollary~\ref{coro:pc_to_pgc} establishes that PGCs subsume
decomposable PCs and in turn, the TPMs subsumed by decomposable PCs.
This raises the question of whether PGCs are \emph{strictly} more expressive
efficient and we give a positive answer:
\begin{thm}
\label{thm:pgc_expressive_efficient}
PGCs are strictly more expressive efficient than decomposable PCs; that is, there exists
a class of probability distributions that can be represented by polynomial-size PGCs
but the sizes of any decomposable PCs that represent the distributions
are at least exponential in the number of random variables.
\end{thm}
We take determinantal point processes (DPPs) as this separating class
of distributions and prove
Theorem~\ref{thm:pgc_expressive_efficient} by showing the following two results:
(1) DPPs, in general, cannot be represented by polynomial-size
decomposable PCs, and (2)~DPPs are tractably subsumed by PGCs. The first result has already
been proved in previous works:

\begin{thm}[\citet{hzhang20,martens2014expressive}]
There exists a class of DPPs such that the the size of any decomposable PCs
that represent them is exponential in the number of random
variables.
\end{thm}

In the next section, we complete this proof by showing that
any DPP can be represented by a PGC of polynomial size in the number
of random variables.


\subsection{PGCs subsume Determinantal Point Processes}
\label{sec:pgc_subsume_dpp}
In this section, we focus on showing that determinantal point processes (DPPs) can be
tractably represented by PGCs. We start by introducing the basics for DPPs.

At a high level, a unique property of DPPs is that they are
tractable representations of probability distributions that express {\it global negative dependence},
which makes them very useful in many applications~\citep{mariet2016kronecker}, such as document and video
summarization~\citep{chao2015large, lin2012learning}, recommender
systems~\citep{zhou2010solving}, and object retrieval~\citep{affandi2014learning}.

In machine learning, DPPs are most often represented by means of an
\emph{L-ensemble}~\citep{borodin2005eynard}:\footnote{Although not every DPP
is an L-ensemble, \citet{MAL-044} show that DPPs that assign non-zero probability
to the empty set (the all-false assignment) are L-ensembles.}
\begin{defn}
\label{def:lensemble}
A probability distribution $\Pr$ over $n$ binary random variables
$\mbf{X} = (X_1, \dots, X_n)$ is an \emph{L-ensemble} if there exists a (symmetric)
positive semidefinite matrix $L \in \R^{n \times n}$ such that for all
$\mbf{x} = (x_1, \dots, x_n) \in \{0, 1\}^{n}$,
\vspace{-2pt}
\begin{align}
\Pr(\mbf{X} = \mbf{x}) \propto \det(L_{\mbf{x}}),
\label{eq:dppkernel}
\end{align}
where $L_{\mbf{x}} = [L_{ij}]_{x_{i} = 1, x_{j} = 1}$ denotes the submatrix of $L$
indexed by those $i,j$ where $x_{i} = 1$ and $x_{j} = 1$. The matrix $L$ is called the
\emph{kernel} for the L-ensemble. To ensure that the distribution is properly normalized, it is necessary to divide
Equation~\ref{eq:dppkernel} by $\det(L + I)$, where $I$ is the $n \times n$
identity matrix~\citep{MAL-044}.
\end{defn}

Consider again the example distribution ${\Pr}_{\beta}$. It is actually a DPP whose kernel
is given by the matrix $L^{\beta}$ in Figure~\ref{fig:example_dist_dpp}.
The probability of the assignment $\mbf{X} = (1, 0, 1)$,
for example, is given by
\vspace{-6pt}
\begin{align*}
  &\Pr(\mbf{X} = (1, 0, 1)) = \frac{\det({L^{\beta}}_{\{1, 3\}})}{\det(L^{\beta} + I)} = \frac{1}{50}
\begin{vmatrix}
 1 &  0 \\
 0 &  4 \\
\end{vmatrix} = 0.08.
\end{align*}

To compute marginal probabilities for L-ensembles, we also need \emph{marginal kernels},
which characterize DPPs in general, as an alternative to L-ensemble kernels.
\begin{defn}
\label{def:dpp}
A probability distribution $\Pr$ is a DPP over $n$ binary random variables
$X_1, \dots, X_n$ if there exists a positive semdidefinite
matrix $K \in \R^{n \times n}$ such that for all $A \subseteq \{1, \dots, n\}$
\vspace{-4pt}
\begin{align}
\label{eq:dpp_marginal_positive}
\Pr(\{X_i = 1\}_{i \in A}) = \det(K_{A}),
\end{align}
where $K_A = [K_{ij}]_{i \in A, j \in A}$ denotes the submatrix of $K$ indexed by elements in
$A$.
\end{defn}
The marginal kernel $K^{\beta}$ for the L-ensemble that represents
the distribution ${\Pr}_{\beta}$ is shown in Figure~\ref{fig:example_dist_dpp}, along
with its kernel $L^{\beta}$.
One can use a generalized version of Equation~\ref{eq:dpp_marginal_positive} to compute
the marginal probabilities $\Pr((X_i \!=\! 1)_{i \in A}, (X_j \!=\! 0)_{j \in B})$ efficiently,
where $A, B \subseteq \{1, \dots, n\}$. We refer to \citet{MAL-044} for further details.

PCs and DPPs support tractable marginals in strikingly different ways,
and we wonder whether these two
tractable languages can be captured by a unified framework.
As mentioned in Section~\ref{sec:pgc_subsume_pc}, it has already been
proved that PCs cannot tractably represent DPPs in general.
We now show that PGCs also tractably subsume DPPs,
providing a positive answer to this open problem.

The key to constructing a PGC representation for DPPs is that
their generating polynomials can be written as determinants over
polynomial rings.
\begin{lem}[\citet{borcea2009negative}]
\label{lem:dpp_generating_polynomial}
The generating polynomial for an L-ensemble with kernel $L$ is given by
\begin{gather}
\label{eq:lensemble_generating_polynomial}
g_L = \frac{1}{\det(L + I)} \det(LZ + I).
\end{gather}
With $Z = \diag(z_1, \dots, z_n)$, the generating polynomial for a DPP with marginal kernel $K$ is given by
\begin{gather}
\label{eq:dpp_generating_polynomial}
g_K = \det(I - K + KZ).
\end{gather}
\end{lem}


Note that the generating polynomials presented in Lemma~\ref{lem:dpp_generating_polynomial}
are just mathematical objects; to use them as tractable models, we need to represent them
in the framework of PGCs.
So let us examine Equations~(\ref{eq:lensemble_generating_polynomial}) and
(\ref{eq:dpp_generating_polynomial}) in detail.
The entries in the matrices $LZ + I$ and $I - K + KZ$ are degree-one
univariate polynomials, which can be easily represented as PGCs.
Thus, to compactly represent DPPs' generating polynomials
as PGCs, we only need to compactly represent the determinant function as a PGC.

There are a variety of polynomial-time division-free algorithms
for computing determinants over rings~\citep{bird2011simple, samuelson1942method,
berkowitz1984computing, mahajan1997combinatorial} and we take
Bird's algorithm~\citep{bird2011simple} as an example.
Bird's algorithm is simply an iteration of certain matrix multiplications and
requires $O(nM(n))$ additions and multiplications, where $M(n)$ is the number
of basic operations needed for a matrix multiplication. We conservatively
assume that $M(n)$ is upper-bounded by $n^3$. Thus when we encode Bird's algorithm as a PGC,
the PGC contains at most $O(n^4)$ sum and product nodes, each with a constant
number of edges. Together with Lemma~\ref{lem:dpp_generating_polynomial}, it
follows that DPPs are tractably subsumed by PGCs.
\begin{thm}
\label{thm:pgc_subsume_dpp}
Any DPP over $n$ binary random variables can be represented by a PGC of size $O(n^4)$.
\end{thm}
We conclude this section with the following remarks:

1. DPPs cannot represent any positive dependence; for example,
$\Pr(X_i = 1, X_j = 1) > \Pr(X_i = 1) \Pr(X_j = 1)$ can never happen for a DPP.
On the other hand, since PGCs are fully general, they are strictly more
expressive than DPPs.

2. In practice, when representing DPPs in the language of
PGCs, we do not need to explicitly construct the sum nodes and
product nodes to form the circuit structures.
Recall from Lemma~\ref{lem:pgc_general_marginal} that marginals are tractable
as long as we can efficiently \emph{evaluate} the PGCs over polynomial rings.
Thus we can simply apply Bird's algorithm, for example, to compute the determinants
from Lemma~\ref{lem:dpp_generating_polynomial}.

3. Since nonsymmetric DPPs~\citep{gartrell2019learning} are defined in the same way
as standard DPPs, except for their kernels $L$ need not be symmetric, they are
also tractably subsumed by PGCs.


\section{Beyond PCs and DPPs}
\label{sec:beyond_pc_dpp}
In the previous section we have demonstrated the expressive power
of PGCs by showing that they are strictly more expressive efficient than
decomposable PCs and DPPs. It is well-known, however, that PCs can use arbitrary
families of tractable distributions at their leaves, including DPPs.
In this section, we construct a simple class of PGCs that
are more interesting than PCs with DPP leaves.

\subsection{Basic Compositional Operations for PGCs}
\label{sec:basic_operation_pgc}
We start by defining the
\emph{sum}, \emph{product} and \emph{hierarchical composition} operations for PGCs.
\begin{prop} \label{prop:pgc_sum_product}
Let $A, B \subset \N^{+}$; denote $\{z_i\}_{i \in A}$ by $\mbf{z}_A$ and
$\{X_i\}_{i \in A}$ by $\mbf{X}_A$.
Let $f(\mbf{z}_A)$ and $g(\mbf{z}_B)$ be the generating polynomials for
distributions $\Pr_f(\mbf{X}_A)$ and $\Pr_g(\mbf{X}_B)$, then, \\
\textbf{Sum}: let $\alpha \in [0, 1]$, then $\alpha f + (1 - \alpha) g$ is
    the generating polynomial for the probability distribution ${\Pr}_{\text{sum}}$ where
    \begin{align*}
    &{\Pr}_{\text{sum}}(\mbf{X}_A = \mbf{a}, \mbf{X}_B = \mbf{b})  \\
    & \quad = \alpha {\Pr}_{f}(\mbf{X}_A = \mbf{a})
    + (1 - \alpha) {\Pr}_{g}(\mbf{X}_B = \mbf{b}).
    \end{align*}
\textbf{Product}: if $A$ and $B$ are disjoint (i.e. $f$ and $g$ depend
    on disjoint sets of variables), then $fg$ is the generating polynomial for the
    probability distribution ${\Pr}_{\text{prod}}$ where
    \begin{align*}
    &{\Pr}_{\text{prod}}(\mbf{X}_A = \mbf{a}, \mbf{X}_B = \mbf{b}) = {\Pr}_{f}(\mbf{X}_A = \mbf{a}){\Pr}_{g}(\mbf{X}_B = \mbf{b}).
    \end{align*}
\end{prop}
\begin{figure}
\begin{center}
\includegraphics[width=0.8\columnwidth]{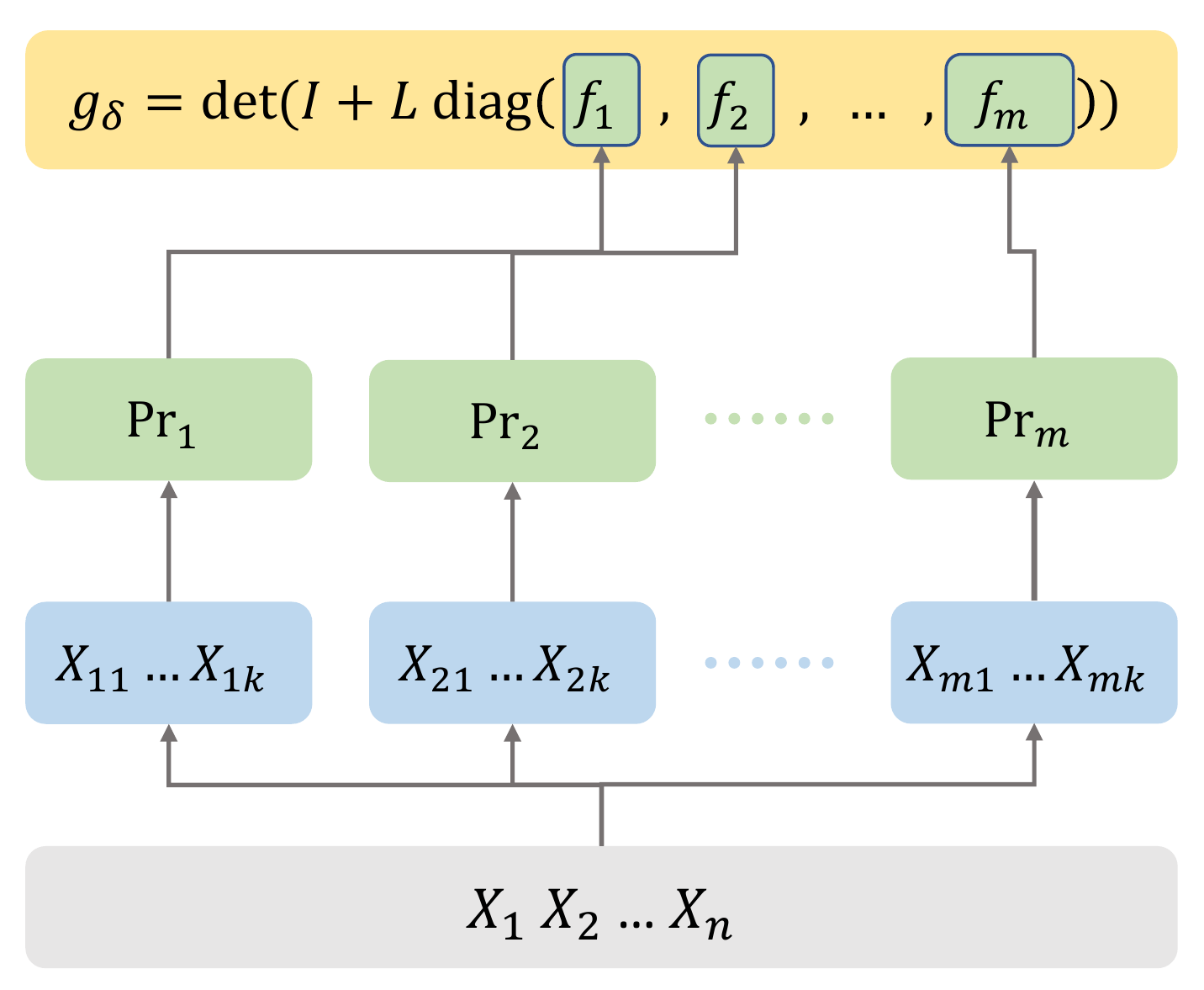}
\end{center}
\caption{An example of the hierarchical composition for PGCs. We partition $n$ binary random variables into $m$ parts, each with $k$ variables. Then, variables
from part $i$ are modeled by the PGC ${\Pr}_i$ with generating polynomial $f_i$.
Let $g_L=\det(I + L\diag(z_1, \dots, z_n))$ be the generating polynomial for a DPP with kernel $L$. Then
$g_{\delta}$ is the hierarchical composition of $g_L$ and $f_i$s. We refer to this architecture
for $g_{\delta}$ as a determinantal PGC.
}
\label{fig:dpp_mixture}
\end{figure}
The sum and product operations described above are basically the same as those for
PCs: the sum distribution ${\Pr}_{\text{sum}}$ is just a mixture over two distributions
${\Pr}_f$ and ${\Pr}_g$, and the product distribution ${\Pr}_{\text{prod}}$ is the point-wise
product of ${\Pr}_f$ and ${\Pr}_g$. The \emph{hierarchical composition} is much more interesting.

\begin{prop}[hierarchical composition]
\label{prop:pgc_composition}
Let ${\Pr}_g$ be a probability distribution with generating polynomial $g(z_1, \dots, z_n)$.
Let $A_1, \dots, A_n$ be disjoint subsets of $\N^{+}$ and $f_1(\mbf{z}_{A_1}), \dots,
f_n(\mbf{z}_{A_n})$ be generating polynomials for ${\Pr}_i$.
We define the hierarchical composition of $g$ and $f_i$s by
$$g_{\comp} = g\eval_{z_i = f_i},$$ which is the generating polynomial obtained by
substituting $z_i$ in $g$ by $f_i$s. In particular, $g_{\comp}$ is a well-defined generating
polynomial that represents a valid probability distribution.
\end{prop}

Unlike the sum and product operations, the hierarchical composition operation
for PGCs does not have an immediate analogue for PCs. This
operation is a simple yet powerful way to construct classes of PGCs;
Figure~\ref{fig:dpp_mixture} shows an example,
which we illustrate in detail in the next section.

\begin{figure*}
\centering
  \begin{subfigure}[b]{0.48\linewidth}
  \centering
  \setlength{\tabcolsep}{2pt}
    {\footnotesize
    \begin{tabular}{c|c c c c c}
    \toprule
       & DPP & Strudel & EiNet & MT & SimplePGC \\ \hline
    nltcs & $-9.23$ & $-6.07$ & $-6.02$ & $\mbf{-6.01}$ & $-6.05^{*}$ \\
    msnbc & $-6.48$ & $\mbf{-6.04}$ & $-6.12$ & $-6.07$ & $-6.06^{\dagger\circ}$ \\
    kdd & $-2.45$ & $-2.14$ & $-2.18$ & $\mbf{-2.13}$ & $-2.14^{*\dagger}$ \\
    plants & $-31.20$ & $-13.22$ & $-13.68$ & $\mbf{-12.95}$ & $-13.52^{\dagger}$ \\
    audio & $-49.31$ & $-42.20$ & $\mbf{-39.88}$ & $-40.08$ & $-40.21^{*}$ \\
    jester & $-63.88$ & $-54.24$ & $\mbf{-52.56}$ & $-53.08$ & $-53.54^{*}$ \\
    netflix & $-64.18$ & $-57.93$ & $\mbf{-56.54}$ & $-56.74$ & $-57.42^{*}$ \\
    accidents & $-35.61$ & $\mbf{-29.05}$ & $-35.59$ & $-29.63$ & $-30.46^{\dagger}$ \\
    retail & $-11.43$ & $\mbf{-10.83}$ & $-10.92$ & $\mbf{-10.83}$ & $-10.84^{\dagger}$ \\
    pumsb & $-51.98$ & $-24.39$ & $-31.95$ & $\mbf{-23.71}$ & $-29.56^{\dagger}$ \\
    dna & $-82.19$ & $-87.15$ & $-96.09$ & $-85.14$ & $\mbf{-80.82}^{*\dagger\circ}$ \\
    kosarek & $-13.35$ & $-10.70$ & $-11.03$ & $\mbf{-10.62}$ & $-10.72^{\dagger}$ \\
    msweb & $-11.31$ & $\mbf{-9.74}$ & $-10.03$ & $-9.85$ & $-9.98^{\dagger}$ \\
    book & $-41.22$ & $-34.49$ & $-34.74$ & $-34.63$ & $\mbf{-34.11}^{*\dagger\circ}$ \\
    movie & $-83.55$ & $-53.72$ & $\mbf{-51.71}$ & $-54.60$ & $-53.15^{*\circ}$ \\
    webkb & $-180.61$ & $\mbf{-154.83}$ & $-157.28$ & $-156.86$ & $-155.23^{\dagger\circ}$ \\
    reuters & $-107.44$ & $-86.35$ & $-87.37$ & $\mbf{-85.90}$ & $-87.65$ \\
    20ng & $-174.43$ & $\mbf{-153.87}$ & $-153.94$ & $-154.24$ & $-154.03^{\circ}$ \\
    bbc & $-278.15$ & $-256.53$ & $\mbf{-248.33}$ & $-261.84$ & $-254.81^{*\circ}$ \\
    ad & $-63.20$ & $-16.52$ & $-26.27$ & $\mbf{-16.02}$ & $-21.65^{\dagger}$ \\
    \bottomrule
    \end{tabular}}
  \caption{Results on the Twenty Datasets benchmark.}
  \label{fig:table_twenty_datasets}
  \end{subfigure}
 ~
  \begin{subfigure}[b]{0.48\linewidth}
  \centering
  \setlength{\tabcolsep}{2pt}
  {\footnotesize
    \begin{tabular}{c|c c c c c}
    \toprule
      & DPP & Strudel & EiNet &  MT & SimplePGC \\ \hline
    apparel & $-9.88$ & $-9.51$ & $-9.24$ & $-9.31$ & $\mbf{-9.10}^{*\dagger\circ}$ \\
  bath & $-8.55$ & $-8.38$ & $-8.49$ & $-8.53$ & $\mbf{-8.29}^{*\dagger\circ}$ \\
  bedding & $-8.65$ & $-8.50$ & $-8.55$ & $-8.59$ & $\mbf{-8.41}^{*\dagger\circ}$ \\
  carseats & $-4.74$ & $-4.79$ & $-4.72$ & $-4.76$ & $\mbf{-4.64}^{*\dagger\circ}$ \\
  diaper & $-10.61$ & $-9.90$ & $-9.86$ & $-9.93$ & $\mbf{-9.72}^{*\dagger\circ}$ \\
  feeding & $-11.86$ & $-11.42$ & $-11.27$ & $-11.30$ & $\mbf{-11.17}^{*\dagger\circ}$ \\
  furniture & $-4.38$ & $-4.39$ & $-4.38$ & $-4.43$ & $\mbf{-4.34}^{*\dagger\circ}$ \\
  gear & $-9.14$ & $-9.15$ & $-9.18$ & $-9.23$ & $\mbf{-9.04}^{*\dagger\circ}$ \\
  gifts & $-3.51$ & $\mbf{-3.39}$ & $-3.42$ & $-3.48$ & $-3.47^{\circ}$ \\
  health & $-7.40$ & $-7.37$ & $-7.47$ & $-7.49$ & $\mbf{-7.24}^{*\dagger\circ}$ \\
  media & $-8.36$ & $\mbf{-7.62}$ & $-7.82$ & $-7.93$ & $-7.69^{\dagger\circ}$ \\
  moms & $-3.55$ & $-3.52$ & $\mbf{-3.48}$ & $-3.54$ & $-3.53^{\circ}$ \\
  safety & $-4.28$ & $-4.43$ & $-4.39$ & $-4.36$ & $\mbf{-4.28}^{*\dagger\circ}$ \\
  strollers & $-5.30$ & $-5.07$ & $-5.07$ & $-5.14$ & $\mbf{-5.00}^{*\dagger\circ}$ \\
  toys & $-8.05$ & $\mbf{-7.61}$ & $-7.84$ & $-7.88$ & $-7.62^{\dagger\circ}$ \\
    \bottomrule
    \end{tabular}}
  \caption{Results on the Amazon Baby Registries benchmark.}
  \label{fig:table_amzn_datasets}
  \end{subfigure}
\caption{Experiment results on the Twenty Datasets and the Amazon Baby Registries, comparing the performance of DPP, Strudel, EiNet, MT
and SimplePGC in terms of average log-likelihood.
Bold numbers indicate the best log-likelihood.
For SimplePGC, annotations $*$, $\dagger$ and $\circ$ mean better 
log-likelihood compared to Strudel, EiNet and MT, respectively.}
\label{fig:table_experiments}
\end{figure*}

\subsection{A Separating Example}
\label{sec:detpgc}
Now we construct a simple class of PGCs that are not trivially subsumed
by PCs with DPPs as leaves.
Figure~\ref{fig:dpp_mixture} gives an outline of its structure.
We construct a model of a probability distribution over $n$ random variables $X_1, \dots, X_n$.
For simplicity we assume $n = mk$ and partition the variables into $m$ parts, each
with $k$ variables. Changing notation, we write $\{X_{11}, \dots, X_{1k}\},
\dots, \{X_{m1}, \dots, X_{mk}\}$. For $1 \leq i \leq m$, let $\Pr_{i}$ be a
PGC over the random variables $X_{i1}, \dots, X_{ik}$ with generating polynomial
$f_i(z_{i1}, \dots, z_{ik})$.
Let $\Pr_{L}$ be a DPP with kernel $L$ and generating polynomial $g_L(z_1, \dots, z_m)$.
Then, the generating polynomial $g_{\delta} = g_{L} \eval_{z_i = f_i}$, namely the hierarchical
composition of $g_{L}$ and $f_i$, defines a PGC ${\Pr}_{\delta}$, which we refer to as
a \emph{determinantal PGC}~(DetPGC).

DPPs are great at modeling negative dependencies but
cannot represent positive dependencies between variables: for the DPP ${\Pr}_{L}$,
${\Pr}_{L}(X_i = 1, X_j = 1) > {\Pr}_{L}(X_i = 1) {\Pr}_{L}(X_j = 1)$ can never happen.
Our construction of DetPGCs aims to equip a DPP model to
capture local positive dependencies. To understand how DetPGCs actually
behave, we compute the marginal probability ${\Pr}_{\delta}(X_{ik} = 1, X_{jl} = 1)$ by
Lemma~\ref{lem:pgc_general_marginal}.

When $X_{ik}$ and $X_{jl}$ belong to the same group (i.e. $i = j$):
\vspace{-2pt}
\begin{align*}
  &{\Pr}_{\delta}(X_{ik} = 1, X_{il} = 1) \\
  &\quad = {\Pr}_{L}(X_i = 1){\Pr}_{i}(X_{ik} = 1, X_{il} = 1);
\end{align*}
that is, when two variables belong to the same group, the dependencies
between them are dominated by $\Pr_i$, giving space for
positive dependencies.

When $X_{ik}$ and $X_{jl}$ belong to different groups (i.e. $i \neq j$):
\begin{align*}
  {\Pr}_{\delta}(X_{ik} = 1, X_{jl} = 1)
\leq {\Pr}_{\delta}(X_{ik} = 1){\Pr}_{\delta}(X_{jl} = 1);
\end{align*}
that is, random variables from different groups
are still negatively dependent, just like variables in the DPP ${\Pr}_{L}$.

We stress that we construct DetPGCs merely to illustrate how the flexibility of PGCs permits us to develop
TPMs that capture structure that is beyond the reach of the standard suite of TPMs, including PCs, DPPs, and
standard combinations thereof. We are \emph{not} proposing DetPGCs
as an ``optimal'' PGC structure for probabilistic modeling. Nevertheless, as we will see, even the simple DetPGC model may be a better model than PCs or DPPs for some kinds of real-world data.

\section{Experiments}
\label{sec:experiments}
This section evaluates PGCs' ability to model real data
on density estimation benchmarks. We use a weighted sum over DetPGCs
as our model. This simple method achieves state-of-the-art performance
on half the benchmarks, illustrating the potential of PGCs in real-world
applications.

\subsection{Datasets}
We evaluate PGCs on two density estimation benchmarks:

1. {\bf Twenty Datasets}~\citep{van2012markov}, which contains 20 real-world
datasets ranging from retail to biology.
These datasets have been used to evaluate various tractable probabilistic
models~\citep{liang2017learning, dang2020strudel, peharz2020einsum}.

2. {\bf Amazon Baby Registries}, which contains 15 datasets,\footnote{
The original benchmark had 17 datasets. We omit the datasets with fewer than 10 variables: decor and pottytrain.
}
each representing a collection of registries or ``baskets'' of baby products
from a specific category such as ``apparel'' and ``bath''. We randomly
split each dataset into train~(70\%), valid~(10\%) and test~(20\%) sets.
This benchmark has been commonly used to evaluate DPP
learners~\citep{gillenwater2014expectation, mariet2015fixed, gartrell2019learning}.

\subsection{Model Structure}
The model we use in our experiments
is a weighted sum over DetPGCs, the example constructed in Section~\ref{sec:detpgc},
which we refer to as SimplePGCs.
Recall from Figure~\ref{fig:dpp_mixture} that a DetPGC is
the hierarchical composition of a DPP and some ``leaf'' PGCs
${\Pr}_1, \dots, {\Pr}_m$. For SimplePGC, we make the simplest choice by setting the ${\Pr}_i$s
to be the fully general PGCs.
In addition, to partition the input variables into $m$ groups
as shown in Figure~\ref{fig:dpp_mixture}, we use
a simple greedy algorithm that aims at putting pairs of positively dependent
variables into the \emph{same} groups.
The structure of a SimplePGC is also governed by two hyperparameters:
the number of DetPGCs in the weighted sum (denoted by $C$) and
the \emph{maximum number of variables} (i.e. $k$ in Figure~\ref{fig:dpp_mixture})
allowed in each group (denoted by $K$).
We tune $C$ and $K$
by a grid search over the following ranges:
$K \in \{1, 2, 5, 7\}$ and $C \in \{1, 4, 7, 10, 20\}$.
Note that our model reduces to a mixture over DPPs when $K=1$.

We implement SimplePGC in PyTorch and learn the parameters by maximum
likelihood estimation (MLE). In particular, we use Adam~\citep{kingma2014adam}
as the optimizing algorithm to minimize the negative log likelihoods given the training sets.
Regularization is done by setting the \emph{weight\_decay} parameter in Adam.
For further details regarding the construction and implementation of
SimplePGCs, please see the Appendix.

\subsection{Baselines}
We compare SimplePGC against four baselines: DPPs, Strudel,
Einsum Networks and Mixture of Trees.

{\bf DPP}: As mentioned, SimplePGC reduces to a mixture over DPPs when
the hyperparameter $K = 1$.
DPPs are learned via SGD. 
We expect PGCs to outperform DPPs on most datasets and be at least as good on all datasets.

{\bf Strudel}:
Strudel~\citep{dang2020strudel} is an
algorithm for learning the circuit structure of structured-decomposable PCs. We include them as
one of the state-of-the-art tractable density estimators. 

{\bf Einsum Networks}: Einsum Networks~\citep{peharz2020einsum} (EiNets) are a deep-learning-style
implementation design for PCs. Compared to Strudel, EiNets are more related to SimplePGC
in the sense that they are both fixed-structure models where only parameters are learned.
The hyper-parameters are chosen by a grid search as suggested by 
\citet{peharz2020einsum}.

{\bf Mixture of Trees} The Mixture of Trees~\citep{meila2000learning} (MT) model is a mixture model
over Chow-Liu trees~\citep{chow1968approximating}. 
MTs are included as a representative of tractable graphical
models with simple yet expressive structures. For learning, we run the 
$mtlearn$ algorithm implemented in the Libra-tk library~\cite{lowd&rooshenas2015}; 
the number of components in MT is chosen by a grid search 
from 2 to 30 with step size 2, as suggested by~\citet{rooshenas2014learning}.  

\subsection{Results and Analysis}
Figure~\ref{fig:table_experiments} shows the experiment results.
We first compare SimplePGC against DPPs.
On both benchmarks, SimplePGC performs significantly better than DPPs on almost
every dataset except for 4 datasets from the Amazon Baby Registries benchmark, where
SimplePGC performs at least as well as DPPs.

Overall, SimplePGC achieves competitive performance when compared
against Strudel, EiNet and MT on both benchmarks. On the Twenty Datasets benchmark,
SimplePGC obtains better average log-likelihood than at least one of the baselines 
(Strudel, EiNet and MT) on 19 out of the 20 datasets and, in particular, 
SimplePGC obtains higher log-likelihood
than \emph{all} of them on 2 datasets. Such results are remarkable,
given the fact that SimplePGC is just a simple hand-crafted PGC architecture with 
little fine-tuning,
while Strudel, EiNet and MT follow from a long line of research aiming to
perform well on exactly the Twenty Datasets benchmark.

The performance of SimplePGC on the Amazon Baby Registries benchmark is
even more impressive: SimplePGC beats \emph{all} of baselines 
on 11 out of 15 datasets and beats at least one of them on all datasets.
One possible reason that SimplePGC performs much better than the other baselines 
on this benchmark is because
these datasets exhibit relatively strong negative 
dependence and SimplePGCs’ DPP-like structure allows them to capture 
negative dependence well.

We also conducted one-sample t-test for the results; for further details please
refer to the Appendix.



\section{PGCs and Strongly Rayleigh Distributions}
\label{sec:pgc_sr_distribution}
At a high level, the study of PCs and graphical models
mainly focuses on constructing
classes of \emph{models} that guarantee tractable exact inference.
A separate line of research in probabilistic machine learning, however,
aims at identifying classes of
\emph{distributions} that support tractable sampling, where generating
polynomials play an essential role.
For example, a well-studied class of distributions are the \emph{strongly Rayleigh (SR) distributions}~\citep{borcea2009negative,li2016fast}, which were first defined in the field of
probability theory for studying negative dependence:

\begin{defn}
A polynomial $f \in \R[z_1, \dots, z_n]$ is \emph{real stable}
if whenever the imaginary part $\text{Im}(z_i) > 0$ for $1 \leq i \leq n$, $f(z_1, \dots, z_n) \neq 0$.
We say that a distribution over $X_1, \dots, X_n$ is
\emph{strongly Rayleigh} (SR) if its generating polynomial is real stable.
\end{defn}

SR distributions contain many important subclasses
such as DPPs and the spanning tree/forest distributions, which have
various applications. From Section~\ref{sec:pgc_subsume_dpp}, we already know that
PGCs can compactly represent DPPs. We now show that PGCs can represent spanning tree distributions in polynomial-size.

We first define the spanning tree distributions. 
Let $G = (V,E)$ be a connected graph with vertex set $V = \{1, \dots, n\}$ and edge set $E$.
Associate to each edge $e \in E$ a variable $z_{e}$ and a weight $w_{e} \in \R_{\geq 0}$.
If $e = \{i, j\}$, let $A_e$ be
the $n \times n$ matrix where $A_{ii} = A_{jj} = 1$, $A_{ij} = A_{ji} = -1$ and all
other entries equal to $0$. Then the \emph{weighted Laplacian} of $G$ is given by
$L(G) = \sum_{e \in E} w_e z_e A_e,$

By the Principal Minors Matrix-Tree Theorem~\citep{chaiken1978matrix},
$$f_G = \det(L(G)_{\setminus \{i\}}) =
  \sum_{T \text{ a spanning tree of G}}
    w^{\text{edges}(T)}
    z^{\text{edges}(T)}$$
is the (un-normalized) generating polynomial for the
spanning tree distribution, and we denote it by ${\Pr}_G$.
Here $L(G)_{\setminus \{i\}}$ denotes the principal
minor of $L(G)$ by removing its $i$th row and column.

As shown in the equation above, ${\Pr}_G$ is supported on the spanning trees of $G$, and the
probability of each spanning tree is proportional to the product of its edge weights.
${\Pr}_G$ is a strongly Rayleigh distribution~\citep{borcea2009negative}, and,
to the best of our knowledge, it is not a DPP unless the edge weights are the same.
By the same argument as in Section~\ref{sec:pgc_subsume_dpp}, we claim that
${\Pr}_G$ can be tractably represented by PGCs.

Thus, we see that there is another natural class of SR distributions -- spanning tree distributions -- that can be represented by PGCs.
More generally, generating
polynomials play a key role in the study of a number of other classes of distributions, including the Ising model~\citep{jerrum1993polynomial},
exponentiated strongly Rayleigh (ESR) distributions~\citep{mariet2018exponentiated} and
strongly log-concave (SLC) distributions~\citep{robinson2019flexible}.
Specifically, most of these distributions are naturally
characterized by their generating polynomials rather than
probability mass functions. This poses a major barrier
to linking them to other probabilistic models. Thus by
showing that PGCs can tractably represent certain
subclasses of SR distributions, we present PGCs as a prospective avenue for bridging this gap.

Although we conjecture that not all SR distributions can be represented by polynomial-size PGCs, we believe that the subclasses of the above distributions that have concise parameterizations should be representable by PGCs. Establishing this for various families is a direction for future~work.

\section{Conclusion and Perspectives}
\label{sec:conclusion}
We conclude by summarizing our contributions and highlighting future research directions.
In this paper, we study the use of probability generating polynomials as a data structure for representing probability distributions.
We showed that their representation as circuits are a TPM, and are strictly more expressive efficient than existing families of TPMs.
Indeed, even a simple example family of distributions that can be represented by PGCs but not PCs or DPPs obtains state-of-the-art performance as a probabilistic model on some datasets.

To facilitate the general use of PGCs for probabilistic modeling, a facinating direction for future work is to build efficient structure learning or `architecture search' algorithms for PGCs.
Theoretically, the main mathematical advantage of generating polynomials was the variety of properties they reveal about a distribution.
This raises the question of whether there are other kinds of useful queries we can support efficiently with PGCs, and where truly lies the boundary of tractable probabilistic inference.


\section*{Acknowledgements}
We thank the reviewers for their detailed and thoughtful feedback and efforts
towards improving this paper. We thank Steven Holtzen, Antonio Vergari and Zhe Zeng
for helpful feedback and discussion.
This work is partially supported by NSF grants \#IIS-1943641, \#IIS-1633857,
\#CCF-1837129, \#CCF-1718380, \#IIS-1908287, and \#IIS-1939677, 
DARPA XAI grant \#N66001-17-2-4032, Sloan and UCLA Samueli Fellowships, 
and gifts from Intel and Facebook Research.

\bibliography{pgc}
\bibliographystyle{icml2021}

\clearpage
\appendix
\onecolumn
\section{Proofs}
\begin{proof}[Proof for Lemma~\ref{lem:pgc_general_marginal}]
We write the generating polynomial for
$\Pr$ as $g(z_1, \dots, z_n) = \sum_{S \subset [n]} \alpha_{S} z^{S}$; then,
\begin{align*}
\Pr(\{X_i = 1\}_{i \in A}, \{X_i = 0\}_{i \in B})) =
\sum_{A \subset S, B \cap S = \emptyset} \alpha_S
\end{align*}
Besides, we also have
\begin{align*}
{\coef}_{|A|}(g\eval_{\asgn}) = \sum \alpha_S {\coef}_{|A|}(z^S\eval_{\asgn}),
\end{align*}
given the assignment
$\asgn := \{\{z_i = t\}_{i \in A}, \{z_i = 0\}_{i \in B}, \{z_i = 1\}_{i \notin {A \cup B}}\}$;
that is, to prove the Proposition,
we only need to show that
${\coef}_{|A|}(z^{S}\eval_{\asgn}) = 1$
for $S \subset [n]$ where $A \subset S$ and $B \cap S = \emptyset$; and
${\coef}_{|A|}(z^{S}\eval_{\asgn}) = 0$ otherwise.

Case 1. Assume $A \subset S$ and $B \cap S = \emptyset$; then
$z^{S}\eval_{\asgn} = t^{|A|}$; hence ${\coef}_{|A|}(z^{S}\eval_{\asgn}) = 1$.

Case 2. Assume $A \not\subset S$ or $B \cap S \neq \emptyset$;
if $B \cap S \neq \emptyset$, then $z^{S}\eval_{\asgn} = 0$; done. Now we
assume $A \not\subset S$. In this case,
$z^{S}\eval_{\asgn} = t^{|S \cap A|}$. It follows from
$S \cap A \subsetneq A$ that $|S \cap A| < |A|$, which implies
that ${\coef}_{|A|}(z^{S}\eval_{\asgn}) = 0$.
\end{proof}

\begin{proof}[Proof for Proposition~\ref{prop:pc_to_pgc}]
Let $\mscr{A}$ be a decomposable and smooth PC. Without loss
of generality we assume $\mscr{A}$ is normalized. For each node
$u$ in $\mscr{A}$, we define $I_{u} = \{i: X_i \text{ or } \ol{X_i} \in \text{scope}(u)\}$
and denote the polynomial that $u$ represents by $m_u$. We first prove the following
intermediate result by a bottom-up induction on $\mscr{A}$:
\begin{align*}
m_u = \sum_{S \subset I_u} \alpha_{S} \prod_{i \in S} X_i \prod_{i \notin S} \ol{X_i},
\end{align*}
where the $\alpha_S$ are some non-negative numbers depending on the node $u$.

Case 1. If $u$ is a leaf node $X_i$ or $\ol{X_i}$, then $m_u$ is
$X_i$ or $\ol{X_i}$; done.

Case 2. If $u$ is a sum node with children $\{v_i\}_{1 \leq i \leq k}$ and
weights $\{w_i\}_{1 \leq i \leq k}$.
$m_u = \sum_{1 \leq i \leq k} w_i m_{v_i}$. By smoothness, $I_{v_i} = I_u$ for all $i$.
Then, by the induction hypothesis:
\begin{align*}
m_u &= \sum_{1 \leq i \leq k} w_i \sum_{S \subset I_{v_i}} \alpha_{iS}
  \prod_{j \in S} X_j \prod_{j \notin S} \ol{X_j} \\
    &= \sum_{1 \leq i \leq k} w_i \sum_{S \subset I_u} \alpha_{iS}
  \prod_{j \in S} X_j \prod_{j \notin S} \ol{X_j} \\
    &= \sum_{S \subset I_u} (\sum_{1 \leq i \leq k} w_i \alpha_{iS})
  \prod_{j \in S} X_j \prod_{j \notin S} \ol{X_j}
\end{align*}

Case 3. If $u$ is a product node with children $\{v_i\}_{1 \leq i \leq k}$. Then,
by decomposability, $I_{v_1}, \dots, I_{v_k}$ are pairwise disjoint; in particular,
for each $S \subset I_u$, $S$ can be uniquely decomposed into
$S_{1} \subset I_{v_1}, \dots, S_{k} \subset I_{v_k}$. Thus,
\begin{align*}
m_u &= \prod_{1 \leq i \leq k} m_{v_i} \\
     &= \prod_{1 \leq i \leq k} \sum_{S_i \subset I_{v_i}} \alpha_{iS}
    \prod_{j \in S_i} X_j \prod_{j \notin S_i} \ol{X_j} \\
     &= \sum_{S_1 \subset I_{v_1}, \dots, S_k \subset I_{v_k}}
    \left(\prod_{1 \leq i \leq k} \alpha_{iS_i}\right)
    \prod_{1 \leq i \leq k} \left(\prod_{j \in S_i} X_j
      \prod_{j \notin S_i} \ol{X_j}\right) \\
     &= \sum_{S \subset I_u} \left(\prod_{1 \leq i \leq k} \alpha_{iS_i}\right)
      \prod_{j \in S} X_j \prod_{j \notin S} \ol{X_j}; \text{\quad with }
        I_u = I_{v_1} \cup \dots \cup I_{v_n} \text{ a disjoint union.}
\end{align*}
Hence the mass polynomial represented by $\mscr{A}$ is given by:
$$m(X_1, \dots, X_n, \ol{X_1}, \dots, \ol{X_n}) = \sum_{S \subset \{1,\dots,n\}} \alpha_{S}
  \prod_{i \in S} X_i \prod_{i \notin S} \ol{X_i}$$
By plugging in $\{X_i = 1\}_{i \in S}, \{X_i = 0\}_{i \notin S}$, it immediately follows
that $\alpha_S = \Pr(\{X_i = 1\}_{i \in S}, \{X_i = 0\}_{i \notin S})$.
\end{proof}

\begin{proof}[Proof for Proposition~\ref{prop:pgc_sum_product}]
Let $A, B \subset \N^{+}$; let $f = \sum_{S \subset A} \beta_S z^S$
and $g = \sum_{S \subset B} \gamma_S z^S$ be the normalized
probability generating polynomials for distributions ${\Pr}_f(\mbf{X}_A)$
and ${\Pr}_g(\mbf{X}_B)$,
respectively.

Case 1 ({\bf Sum}). First, we view $f$ and $g$ as polynomials over $\{z_i\}_{i \in A \cup B}$
by setting $\beta_S = 0$ $\forall S \not\subset A$, and $\gamma_S = 0$
$\forall S \not\subset B$, which is equivalent to,
from the perspective of probability distributions,
viewing ${\Pr}_f$ and ${\Pr}_g$ as distributions
over $\mbf{X}_{A \cup B}$ such that
\begin{align*}
    {\Pr}_f(\mbf{X}_A = \mbf{a}, \mbf{X}_B = \mbf{b}) =
\begin{cases}
    {\Pr}_f(\mbf{X}_A = \mbf{a}), & \text{if } b_i = 0 \text{ for } i \in (A \cup B) \backslash A\\
    0,              & \text{otherwise}
\end{cases}
\end{align*}
and
\begin{align*}
    {\Pr}_g(\mbf{X}_A = \mbf{a}, \mbf{X}_B = \mbf{b}) =
\begin{cases}
    {\Pr}_g(\mbf{X}_B = \mbf{b}), & \text{if } a_i = 0 \text{ for } i \in (A \cup B) \backslash B\\
    0,              & \text{otherwise}
\end{cases}
\end{align*}
Then,
\begin{align*}
\alpha f + (1 - \alpha) g
&= \alpha \sum_{S \subset A} \beta_S z^S + (1 - \alpha) \sum_{S \subset B} \gamma_S z^S \\
&= \sum_{S \subset A \cup B} \left(\alpha \beta_S + (1 - \alpha)\gamma_S\right) z^S,
\end{align*}
where $\alpha \beta_S + (1 - \alpha) \gamma_S$ are clearly non-negative, and
$\sum_{S \subset A \cup B} \alpha \beta_S + (1 - \alpha) \gamma_S =
\alpha \sum_{S \subset A \cup B} \beta_S +
(1 - \alpha) \sum_{S \subset A \cup B} \gamma_S = \alpha + (1 - \alpha) = 1$. That is,
$\alpha f + (1 - \alpha) g$ is a valid probability generating polynomial for a distribution,
${\Pr}_{sum}$.

For assignments
$\mbf{X}_A = \mbf{a}$ and $\mbf{X}_B = \mbf{b}$ with no conflict ($A$ and $B$ are
not necessarily disjoint),
let $S = \{i \in A: a_i = 1\} \cup \{i \in B: b_i = 1\}$. By definition,
${\Pr}_{sum}(\mbf{X}_A = \mbf{a}, \mbf{X}_B = \mbf{b})$ is given by the coefficient
of the term $z^{S}$, which is
\begin{align*}
  &\alpha \beta_{S} + (1 - \alpha) \gamma_{S} \\
= &\alpha {\Pr}_f(\mbf{X}_A = \mbf{a}, \mbf{X}_B = \mbf{b}) +
  (1 - \alpha) {\Pr}_g(\mbf{X}_A = \mbf{a}, \mbf{X}_B = \mbf{b})\\
= &\alpha {\Pr}_f(\mbf{X}_A = \mbf{a}) +
  (1 - \alpha) {\Pr}_g(\mbf{X}_B = \mbf{b})  \text{ for short.}
\end{align*}

Case 2 ({\bf Product}). We assume $A \cap B = \emptyset$. Then,
\begin{align*}
fg &= \left(\sum_{S \subset A} \beta_S z^S\right)
      \left(\sum_{T \subset B} \gamma_T z^T\right) \\
  &= \sum_{S \subset A, T \subset B} \beta_{S} \gamma_{T} z^{S}z^{T}
\end{align*}
As $A$ and $B$ are disjoint, $z^{S}z^{T}$ are multiaffine.
On top of that, $\beta_{S} \gamma_{T} \geq 0$ and
$\sum_{S \subset A, T \subset B} \beta_{S} \gamma_{T}
= \sum_{S \subset A} \beta_{S} \sum_{T \subset B} \gamma_{T} = 1$. Thus,
$fg$ is a valid probability generating polynomial for a distribution, ${\Pr}_{prod}$.

For assignments $\mbf{X}_A = \mbf{a}, \mbf{X}_B = \mbf{b}$, we set
$S_{\mbf{a}} = \{i \in A: a_i = 1\}$ and $S_{\mbf{b}} = \{i \in B: b_i = 1\}$.
Then, by definition, ${\Pr}_{prod}(\mbf{X}_A = \mbf{a}, \mbf{X}_B = \mbf{b})$ is
given by the coefficient of the term $z^{S_{\mbf{a}} \cup S_{\mbf{b}}}$, which is
$\beta_{S_{\mbf{a}}} \gamma_{S_{\mbf{b}}}=
{\Pr}_{f}(\mbf{X}_A = \mbf{a}){\Pr}_{f}(\mbf{X}_B = \mbf{b})$.

\end{proof}

\begin{proof}[Proof for Proposition~\ref{prop:pgc_composition}]
Let $g(z_1, \dots, z_n)$ be a normalized
probability generating polynomial. Let $A_1, \dots, A_n$ be disjoint subsets
of $\N^{+}$ and $f_1(\mbf{z}_{A_1}), \dots, f_n(\mbf{z}_{A_n})$ be normalized generating
polynomials. Write $g = \sum_{S \subset \{1, \dots, n\}} \alpha_S z^S$. Then,
$$g\eval_{z_i = f_i} = \sum_{S \subset \{1, \dots, n\}} \alpha_S \prod_{i \in S} f_i$$
It follows from Proposition~\ref{prop:pgc_sum_product} (product operation)
that $\prod_{i \in S} f_i$ are valid generating polynomials for
$S \subset \{1, \dots, n\}$; again by Proposition~\ref{prop:pgc_sum_product}
(sum operation), $g\eval_{z_i = f_i} = \sum_{S \subset \{1, \dots, n\}} \alpha_S \prod_{i \in S} f_i$ is
a valid generating polynomial.
\end{proof}

\section{Experiments}
\label{appendix:experiments}
\subsection{The Construction of SimplePGC}
A SimplePGC is a weighted sum over several DetPGCs, which are defined in
Section~\ref{sec:detpgc}. The structure of a SimplePGC is governed by
two hyperparameters, the number of DetPGCs in the weighted sum (denoted by $C$),
and the maximum number of the variables (i.e. $k$ in Figure~\ref{fig:dpp_mixture})
in the leaf distributions of the DetPGCs (denoted by $K$).

{\bf Partitioning Variables} To construct SimplePGC,
we first partition the variables $X_1, \dots, X_n$ into several groups.
The idea is, as shown in Section~\ref{sec:detpgc},
for a DetPGC, variables from different groups have to be negatively dependent, so
we want to put pairs of variables that are positively dependent in the same group.
Given some training examples $D_1, \dots, D_l$,
we estimate the probabilities $\Pr(X_i = 1)$ and $\Pr(X_i = 1, X_j = 1)$
by counting; in particular, we set
$$\Pr(\text{event}) = \frac{|{D_i \text{where event is true}}|}{l}.$$
Then, inspired
by the definition of \emph{pairwise mutual information},
when $\Pr(X_i = 1, X_j = 1) > 0$, we use the quantity
\begin{align}
\label{eq:edge_weight}
w_{ij} = \Pr(X_i = 1, X_j = 1) \log\frac{\Pr(X_i = 1, X_j = 1)}{\Pr(X_i = 1)\Pr(X_j = 1)},
\end{align}
to measure the
degree of positive dependence between $X_i$ and $X_j$. Note that $X_i$ and $X_j$ are positively
dependent if $w_{ij} > 0$.
Then we partition the variables into groups by the following greedy algorithm.
\begin{algorithm}[H]
   \caption{Partition Variables}
   \label{algo:partition_variables}
\begin{algorithmic}
\STATE {\bfseries Input:} variables $\{X_i\}_{1 \leq i \leq n}$,
weights $\{w_{ij}\}_{1\leq i < j \leq n}$ as defined by Equation~(\ref{eq:edge_weight})
\STATE {\bfseries Output:} function $group$ that maps $X_i$ to the group
that $X_i$ belongs to
\STATE {\bfseries Initialization:} $group(X_i) \leftarrow \{X_i\}$, $W \leftarrow \{w_{ij} > 0\}_{1 \leq i < j \leq n}$
\STATE
\STATE sort $W$ in descending order
\FOR{$w_{ij}$ {\bfseries in} $W$}
\STATE $union \leftarrow group(X_i) \cup group(X_j)$
\IF{$|union| \leq K$}
\STATE $group(X_i) \leftarrow union$
\STATE $group(X_j) \leftarrow union$
\ENDIF
\ENDFOR
\end{algorithmic}
\end{algorithm}

{\bf Leaf PGCs} After we partition the variables in to groups, we feed
them to the leaf PGCs ${\Pr}_{i}$, as shown in Figure~\ref{fig:dpp_mixture}.
${\Pr}_i$ can be any PGCs;
for SimplePGCs, we make the simplest choice by setting them to be the fully
general distributions. For the leaf distribution ${\Pr}_i$ over variables
$\{X_{i1}, \dots, X_{ik}\}$, we let
$$g_i = \frac{1}{Z_i}
{\sum}_{\emptyset \neq S \subset \{i1, \dots, ik\}} \exp(\theta_{i,S}) z^{S}$$
be its generating polynomial with parameters
${\theta_{i, S}}$ and normalizing constant $Z_i$.
${\Pr}_i$s are fully general except for the constraint
that all-zero assignment must have zero probability; this constraint is
nothing but a trick that makes implementation easier.

\subsection{Supplementary Experiment Results}
In addition to the experiment results presented in Figure~\ref{fig:table_experiments},
we also conducted one-sample t-tests.
The results of the statistical
test are shown in Figure~\ref{fig:table_experiments_t_test}. For the Twenty Datasets
benchmark, the log-likelihood of Strudel is only statistically better
than SimplePGC on 6 out of 20 datasets; the log-likelihood of EiNets is only
statistically better than SimplePGC on 2 out of 20 datasets;
the log-likelihood of MT is statistically better than
SimplePGC on 7 out of 20 datasets.
For the Amazon Baby Registries benchmark, the log-likelihoods of Strudel, EiNets and MT
are statistically better than SimplePGC on none of the datasets.

It would also help if we estimate the confidence intervals of the average test log-likelihoods via cross validation. However, since we have four baselines and not all of their learning algorithms were designed to be fast, the computation cost for estimating the confidence intervals would be infeasible.

\begin{figure}[h]
\centering  
  \begin{subfigure}[b]{1.0\linewidth}  
  \centering  
  {\footnotesize
    \begin{tabular}{c|c c c c c c c}
    \toprule
       & Strudel & EiNet & MT & SimplePGC & vs. Strudel & vs. EiNet & vs. MT\\ \hline
    nltcs & $-6.07$ & $-6.02$ & $\mbf{-6.01}$ & $-6.05$ & $=$ & $=$ & $=$ \\
    msnbc & $\mbf{-6.04}$ & $-6.12$ & $-6.07$ & $-6.06$ & $<$ & $>$ & $=$ \\
    kdd & $-2.14$ & $-2.18$ & $\mbf{-2.13}$ & $-2.14$ & $=$ & $>$ & $=$ \\
    plants & $-13.22$ & $-13.68$ & $\mbf{-12.95}$ & $-13.52$ & $<$ & $=$ & $<$ \\
    audio & $-42.20$ & $\mbf{-39.88}$ & $-40.08$ & $-40.21$ & $>$ & $=$ & $=$ \\
    jester & $-54.24$ & $\mbf{-52.56}$ & $-53.08$ & $-53.54$ & $>$ & $<$ & $<$ \\
    netflix & $-57.93$ & $\mbf{-56.54}$ & $-56.74$ & $-57.42$ & $>$ & $<$ & $<$ \\
    accidents & $\mbf{-29.05}$ & $-35.59$ & $-29.63$ & $-30.46$ & $<$ & $>$ & $<$ \\
    retail & $\mbf{-10.83}$ & $-10.92$ & $\mbf{-10.83}$ & $-10.84$ & $=$ & $=$ & $=$ \\
    pumsb & $-24.39$ & $-31.95$ & $\mbf{-23.71}$ & $-29.56$ & $<$ & $>$ & $<$ \\
    dna & $-87.15$ & $-96.09$ & $-85.14$ & $\mbf{-80.82}$ & $>$ & $>$ & $>$ \\
    kosarek & $-10.70$ & $-11.03$ & $\mbf{-10.62}$ & $-10.72$ & $=$ & $=$ & $=$ \\
    msweb & $\mbf{-9.74}$ & $-10.03$ & $-9.85$ & $-9.98$ & $<$ & $=$ & $=$ \\
    book & $-34.49$ & $-34.74$ & $-34.63$ & $\mbf{-34.11}$ & $=$ & $=$ & $=$ \\
    movie & $-53.72$ & $\mbf{-51.71}$ & $-54.60$ & $-53.15$ & $=$ & $=$ & $=$ \\
    webkb & $\mbf{-154.83}$ & $-157.28$ & $-156.86$ & $-155.23$ & $=$ & $=$ & $=$ \\
    reuters & $-86.35$ & $-87.37$ & $\mbf{-85.90}$ & $-87.65$ & $=$ & $=$ & $<$ \\
    20ng & $\mbf{-153.87}$ & $-153.94$ & $-154.24$ & $-154.03$ & $=$ & $=$ & $=$ \\
    bbc & $-256.53$ & $\mbf{-248.33}$ & $-261.84$ & $-254.81$ & $=$ & $=$ & $=$ \\
    ad & $-16.52$ & $-26.27$ & $\mbf{-16.02}$ & $-21.65$ & $<$ & $>$ & $<$ \\
    \bottomrule
    \end{tabular}}
  \caption{One-sided t-test results on the Twenty Datasets benchmark.}
  \label{fig:table_twenty_datasets_t_test}
  \end{subfigure}
  \\
  \begin{subfigure}[b]{1.0\linewidth}
  \centering
  {\footnotesize
    \begin{tabular}{c|c c c c c c c}
    \toprule
      & Strudel & EiNet & MT & SimplePGC & vs. Strudel & vs. EiNet & vs. MT \\ \hline
    apparel & $-9.51$ & $-9.24$ & $-9.31$ & $\mbf{-9.10}$ & $>$ & $=$ & $>$ \\
    bath & $-8.38$ & $-8.49$ & $-8.53$ & $\mbf{-8.29}$ & $=$ & $>$ & $>$ \\
    bedding & $-8.50$ & $-8.55$ & $-8.59$ & $\mbf{-8.41}$ & $=$ & $=$ & $>$ \\
    carseats & $-4.79$ & $-4.72$ & $-4.76$ & $\mbf{-4.64}$ & $>$ & $=$ & $>$ \\
    diaper & $-9.90$ & $-9.86$ & $-9.93$ & $\mbf{-9.72}$ & $>$ & $=$ & $>$ \\
    feeding & $-11.42$ & $-11.27$ & $-11.30$ & $\mbf{-11.17}$ & $>$ & $=$ & $=$ \\
    furniture & $-4.39$ & $-4.38$ & $-4.43$ & $\mbf{-4.34}$ & $=$ & $=$ & $=$ \\
    gear & $-9.15$ & $-9.18$ & $-9.23$ & $\mbf{-9.04}$ & $=$ & $>$ & $>$ \\
    gifts & $\mbf{-3.39}$ & $-3.42$ & $-3.48$ & $-3.47$ & $=$ & $=$ & $=$ \\
    health & $-7.37$ & $-7.47$ & $-7.49$ & $\mbf{-7.24}$ & $>$ & $>$ & $>$ \\
    media & $\mbf{-7.62}$ & $-7.82$ & $-7.93$ & $-7.69$ & $=$ & $=$ & $=$ \\
    moms & $-3.52$ & $\mbf{-3.48}$ & $-3.54$ & $-3.53$ & $=$ & $=$ & $=$ \\
    safety & $-4.43$ & $-4.39$ & $-4.36$ & $\mbf{-4.28}$ & $>$ & $>$ & $=$ \\
    strollers & $-5.07$ & $-5.07$ & $-5.14$ & $\mbf{-5.00}$ & $=$ & $=$ & $>$ \\
    toys & $\mbf{-7.61}$ & $-7.84$ & $-7.88$ & $-7.62$ & $=$ & $>$ & $>$ \\
    \bottomrule
    \end{tabular}}
  \caption{One-sided t-test results on the Amazon Baby Registries benchmark.}
  \label{fig:table_amzn_datasets_t_test}
  \end{subfigure}
\caption{Results for one-sample two-sided t-test on two benchmarks with $p = 0.1$.
In the 5th and 6th columns of the tables, $=$ means a statistical tie, $>$ means
that the log-likelihood of SimplePGC is statistically better, and $<$ means that
of SimplePGC is statistically worse.
Note that a statistical tie
does not necessarily mean there is no difference in terms of performance.
Bold numbers indicate the best log-likelihood.}
\label{fig:table_experiments_t_test}
\end{figure}

\end{document}